\documentclass[accepted,specialissue]{melba}

\usepackage{mwe} 

\usepackage{amsmath,amsfonts}

\usepackage{adjustbox} 
\usepackage{pifont} 
\usepackage[table]{xcolor} 

\newcommand{\argmin}{\mathop{\mathrm{arg\,min}}}
\newcommand{\cmark}{\ding{51}}%
\newcommand{\xmark}{\ding{55}}%

\melbaid{2026:005}  
\doi{10.59275/j.melba.2026-8b23}
\melbaauthors{Weiherer \textit{et al.}}  
\email{maximilian.weiherer@fau.de}
\volume{2026}
\firstpageno{95}  
\melbayear{2026}  
\datesubmitted{07/2025}  
\datepublished{02/2026}  

\melbaspecialissue{MELBA–BVM 2025 Special Issue}
\melbaspecialissueeditors{Andreas Maier, Thomas Deserno, Heinz Handels, Klaus Maier-Hein, Christoph Palm, Thomas Tolxdorff, Katharina Breininger}

\ShortHeadings{Learning Neural Parametric 3D Breast Shape Models}{Weiherer \textit{et al.}}

\title{Learning Neural Parametric 3D Breast Shape Models for Metrical Surface Reconstruction From Monocular RGB Videos}

\author{
	\firstname Maximilian \surname Weiherer\aff{1,2},
	\firstname Antonia \surname von Riedheim\aff{3},
    \firstname Vanessa \surname Brébant\aff{3},
    \firstname Bernhard \surname Egger\aff{1}$^*$,
    \firstname Christoph \surname Palm\aff{2}$^*$
}
\affiliations{
	\num 1 \addr Visual Computing Erlangen, Friedrich-Alexander-Universtität Erlangen-Nürnberg, Erlangen, Germany \\
	\num 2 \addr Regensburg Medical Image Computing (ReMIC), OTH Regensburg, Regensburg, Germany \\
	\num 3 \addr Department for Plastic, Hand and Reconstructive Surgery, University Hospital Regensburg, Regensburg, Germany\\
    * \it{equal contribution}
}

\abstract{
    We present a neural parametric 3D breast shape model and, based on this model, introduce a low-cost and accessible 3D surface reconstruction pipeline capable of recovering accurate breast geometry from a monocular RGB video.
    In contrast to widely used, commercially available yet expensive 3D breast scanning solutions \textit{and} existing low-cost alternatives, our method requires neither specialized hardware nor proprietary software and can be used with any device that is able to record RGB videos.
    The key building blocks of our pipeline are a state-of-the-art, off-the-shelf Structure-from-Motion pipeline, paired with a parametric breast model for robust surface reconstruction.
    Our model, similarly to the recently proposed implicit Regensburg Breast Shape Model (iRBSM), leverages implicit neural representations to model breast shapes.
    However, unlike the iRBSM, which employs a single \textit{global} neural Signed Distance Function (SDF), our approach---inspired by recent state-of-the-art face models---decomposes the implicit breast domain into multiple smaller regions, each represented by a \textit{local} neural SDF anchored at anatomical landmark positions.
    When incorporated into our surface reconstruction pipeline, the proposed model, dubbed liRBSM (short for \textit{localized} iRBSM), significantly outperforms the iRBSM in terms of reconstruction quality, yielding more detailed surface reconstruction than its global counterpart. 
    Overall, we find that the introduced pipeline is able to recover high-quality and metrically correct 3D breast geometry within an error margin of less than 2 mm.
    Our method is fast (requires less than six minutes), fully transparent and open-source, and together with the model publicly available at \url{https://rbsm.re-mic.de/local-implicit}.
}


\begin{document}

\twocolumn[\maketitle]


\begin{figure*}[t]
	\centering
	\includegraphics[width=\linewidth]{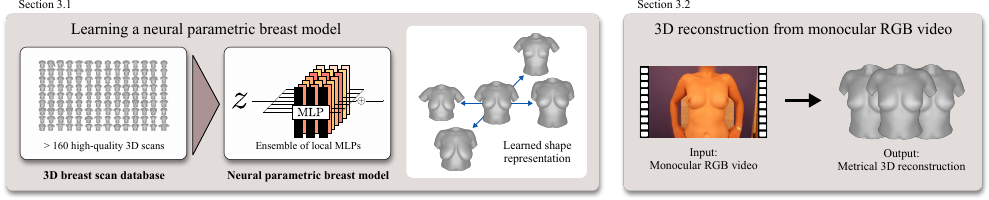}
	\caption{\textbf{Contributions.} We propose a 3D parametric breast model along with a model-based surface reconstruction pipeline that is able to accurately reconstruct 3D breast surfaces from a single monocular RGB video. Our model, trained on over 160 breast scans, builds upon a \textit{localized} neural implicit representation and represents breast geometry as an ensemble of local MLPs instead of a single global network, significantly improving the level of detail.}
    \label{fig:contributions}
\end{figure*}

\section{Introduction}
\enluminure{D}{espite} recent advances and emerging applications, research evolving around 3D parametric models for the female breast remains rare.
Since the seminal work of \citet{seo2007breast}, who developed the first classical mesh-based statistical breast shape model, the field has seen relatively little progress---especially when compared to the face domain \citep{egger2020face}---likely due to the sensitivity of the data and the resulting lack of publicly available 3D breast scan datasets.
Nevertheless, over the years, 3D parametric breast models have found use in a range of medical applications, including automatic breast volume estimation from surface scans \citep{seo2007breast,goepper2020breast}, plastic surgery simulation \citep{kim2008breast}, model-based breast segmentation in Magnetic Resonance Imaging (MRI) volumes \citep{gallego2011breast}, and surface reconstruction from 2D photographs \citep{ruiz2018breast} or point clouds obtained using depth cameras \citep{mazier2021rigged}.
Potential future use cases beyond the clinical domain include virtual try-on and custom bra design. 
A common limitation shared by \textit{all} of these approaches is the absence of publicly available code and trained models, which prevents broader adoption and hinders future research.
As a consequence and as of today, none of the above-mentioned methods is effectively accessible to clinicians or other researchers, and only two publicly available 3D breast models exist: the Regensburg Breast Shape Model (RBSM; \cite{weiherer2023rbsm}) and the recently proposed state-of-the-art \textit{implicit} RBSM (iRBSM; \cite{weiherer2025irbsm}). 
While the RBSM is a traditional mesh-based statistical shape model built by applying Principal Component Analysis (PCA) to a set of 110 non-rigidly registered 3D breast scans, the iRBSM takes a fundamentally different approach by leveraging implicit neural representations to learn a realistic and expressive parametric breast model from over 160 subjects.
Instead of using triangular meshes as surface representation, the iRBSM represents breast shapes as the zero-level set of a neural Signed Distance Function (SDF), modeled with a simple coordinate-based Multi-Layer Perceptron (MLP).
Once trained, triangular meshes can be extracted from the implicit volume using marching cubes algorithm \citep{lorensen1987mc}.
A major advantage of implicit representations over PCA-based models is that they eliminate the need for training data to be in correspondence, entirely removing the reliance on computationally demanding and error-prone non-rigid surface registration---a task that is particularly difficult in feature-less and partially occluded breast shapes as discussed in \citep{weiherer2025irbsm}.

In this work, we build upon the neural implicit representation of the iRBSM and propose a novel open-access 3D parametric breast model.
Our model is inspired by recent state-of-the-art face models \citep{zheng2022imface,giebenhain2023nphm,potamias2025imhead} and employs a decomposition of the implicit breast domain into multiple smaller regions, each of which is modeled using a \textit{local} neural SDF anchored at anatomical landmark positions instead of a single global network.
Compared to the iRBSM, our model provides a significantly higher level of detail, enabling the recovery of fine anatomical structures such as skin folds and nipples.
Due to its local nature, we refer to our model as liRBSM (short for \textit{localized} iRBSM).

Largely independent from these developments---with the exception of \citep{ruiz2018breast,mazier2021rigged}---3D breast scanning (or imaging) has become standard practice in plastic surgery over the past two decades, often used to digitally assess breast volume \citep{kovacs2007breast,lee2016breast,seoud2017breast,gouveia2021breast} or symmetry \citep{eder2012breast,brebant2022breast,noisser2022breast,bai2023breast}, perform anthropometric measurements \citep{hartmann2021breast,leusink2021breast,wang2025breast}, or simulate surgical outcomes \citep{kim2008breast,georgii2014breast}.
To create surface scans of the breast, most researchers and clinicians rely on commercially available 3D scanning solutions, typically requiring special hardware and proprietary software, such as Artec's handheld Eva\footnote{\url{https://www.artec3d.com/portable-3d-scanners/artec-eva}} or Canfield's portable Vectra H2\footnote{\url{https://www.canfieldsci.com/imaging-systems/vectra-h2-body}} or static Vectra XT\footnote{\url{https://www.canfieldsci.com/imaging-systems/vectra-xt-3d-imaging-system}} system.
Although established, well validated for numerous breast-related applications, and known to produce highly accurate 3D breast scans, due to their high purchase price (currently $\sim$10,000 Euros for Artec Eva; around 20,000 Dollars for Vectra H2), these systems are typically only available to large hospitals or clinical facilities.
As a result, a recent line of research started to investigate various cost-effective alternatives based on \textit{consumer-grade} hardware, most notably the iPhone \citep{pinto2022breast,han2023breast,rudy2024breast,behrens2024breast,dijkman2024breast,kyriazidis2025breast,chrobot2025breast}, in combination with freely available software such as the 3D Scanner App\footnote{\url{https://www.3dscannerapp.com}}.
While these methods offer more accessibility than Artec's or Canfield's systems, they are still limited to certain devices and rely on closed-source software, restricting transparency and full control over the surface reconstruction process which ultimately raises privacy concerns and imposes laborious workflows \citep{chrobot2025breast}.

To tackle these shortcomings and based on the proposed parametric breast model, we introduce a new low-cost, accessible, and accurate 3D breast surface reconstruction method that only requires a monocular RGB video as input.
Our method produces accurate metrical 3D reconstructions, is fast (takes less than six minutes to run on a standard computer with a consumer-grade graphics card), well-documented, and fully open-source.
Moreover, it can be used in combination with any capturing device that is able to record RGB videos.
From a technical perspective, our pipeline combines a recent state-of-the-art and off-the-shelf Structure-from-Motion pipeline with robust model-based surface reconstruction.

Our key contributions are outlined in Figure \ref{fig:contributions} and can be summarized as follows:
\begin{itemize}
    \item We present a new 3D parametric breast model that builds upon a \textit{localized} neural implicit representation, significantly increasing the level of detail.
    \item We propose a low-cost and accessible 3D surface reconstruction pipeline that is able to accurately recover metrical breast surfaces from just a single monocular RGB video, captured using commodity hardware. 
    \item To encourage further research and widespread adoption, we publicly release our model and surface reconstruction pipeline along with an easy-to-use graphical user interface that runs on all common operating systems and (optionally) without a graphics card. Both available at \url{https://rbsm.re-mic.de/local-implicit}.
\end{itemize}


\section{Related Work}
We begin with a comprehensive overview of existing parametric 3D breast shape models, followed by a review of related work on 3D breast surface reconstruction methods. 
To contextualize our work within the broad landscape of shape modeling and (model-based) surface reconstruction from RGB images, we finally briefly summarize related literature on full-body models and reconstruction techniques.

\subsection{Parametric 3D Breast Models}
With the exception of \citep{gallego2011breast}, all existing models have been trained on 3D breast scans acquired in a standing position.

\cite{seo2007breast} were the first to build a parametric breast model from 28 scans to analyze breast volume and surface measurements.
They assumed symmetric breasts by simply mirroring the right breast.

\cite{gallego2011breast} learned a PCA-based model from 415 MRI-extracted 3D breast surfaces captured in \textit{prone} position.
Their model is specifically tailored for automatic, model-based breast segmentation in MRI volumes; hence, it was constructed using data only from a single breast rather than the full thoracic region.

\cite{ruiz2018breast} built a 3D breast model from 310 scans and subsequently fit their model onto 3D breast scans or a set of three unconstrained 2D photographs taken from frontal and lateral views ($\pm$ 90 degrees).

\cite{mazier2021rigged} proposed a \textit{rigged} 3D breast model built from 55 artist-created blendshapes to transfer surgical reference patterns drawn on the model's mean shape onto any patient in any position by non-rigidly registering the model to a patient's 3D breast scan.
Due to the synthetic nature of the data, their model is likely to generate rather unnatural-looking breast shapes, and the overall accuracy measured in terms of mean absolute error and landmark error of the surface registrations is moderate.

\cite{weiherer2023rbsm} published the first publicly available PCA-based model of the female breast, trained on 110 high-quality 3D breast scans.
While representing a notable advancement towards making 3D parametric breast models accessible, their model exhibits correspondence errors as a result of occluded underbusts in large or sagged breasts---a common problem in classical PCA-based \textit{breast} models built from surface-only 3D breast scans \citep{seo2007breast}.

To account for this, \cite{weiherer2025irbsm} recently introduced the iRBSM, the first \textit{implicit} 3D breast shape model. 
Based on correspondence-free neural implicit representations (representing breast shapes as the zero-level set of implicit surfaces parametrized with an MLP) and trained on 168 breast scans, their model is able to generate diverse yet plausible breast shapes. 
However, when fitted onto point clouds, the global nature of the iRBSM misses fine details such as skin folds and nipples. 

Only the RBSM \citep{weiherer2023rbsm} and iRBSM \citep{weiherer2025irbsm} are publicly available.
A compact summary of all models is given in Table \ref{tab:overview_breast_models}.

\begin{table}
    \begin{adjustbox}{width=\linewidth}
        \begin{tabular}[t]{lllllcc}
        \toprule
        Study & Representation & Method & Training Data & Pose & \#Subjects & Public? \\
        \midrule
        \cite{seo2007breast} & Mesh-based & PCA & 3D breast scans & Standing & 28 & \xmark \\
        \cite{gallego2011breast} & Mesh-based & PCA & Segmented MRIs & Prone & 415 & \xmark \\
        \cite{ruiz2018breast} & Mesh-based & PCA & 3D breast scans & Standing & 310 & \xmark \\
        \cite{mazier2021rigged} & Mesh-based & PCA & Blendshapes & Standing & 55 & \xmark \\
        \cite{weiherer2023rbsm} & Mesh-based & PCA & 3D breast scans & Standing & 110 & \cmark \\
        \cite{weiherer2025irbsm} & Implicit (Global) & Deep & 3D breast scans & Standing & 168 & \cmark \\
        \midrule
        Ours & Implicit (Local) & Deep & 3D breast scans & Standing & 168 & \cmark \\
        \bottomrule
        \end{tabular}
    \end{adjustbox}
    \caption{\textbf{Overview of existing 3D parametric breast models.} Our model is the first to use a \textit{localized} neural implicit representation.}
    \label{tab:overview_breast_models}
\end{table}

\begin{table*}[t]
    \begin{adjustbox}{width=\textwidth}
        \begin{tabular}[t]{lllllccc}
        \toprule
        Study & Input Data & Capturing Device & Software (SW) & Compatible With & Handheld? & Public SW? & Open-source SW? \\
        \midrule
        \cite{costa2014breast} & RGB-D & Microsoft Kinect & Kinect SDK & Any RGB-D camera$^*$ & \xmark & \cmark & \xmark \\
        \cite{henseler2014breast} & RGB-D & Microsoft Kinect & Kinect SDK + Custom & Any RGB-D camera$^*$ & \xmark & \xmark & \xmark \\
        \cite{poehlmann2014breast} & RGB-D & Microsoft Kinect & Kinect SDK & Any RGB-D camera$^*$ & \xmark & \cmark & \xmark \\
        \cite{wheat2014breast} & RGB-D & 2 Microsoft Kinects & Kinect SDK + Custom & Any RGB-D camera$^*$ & \xmark & \xmark & \xmark \\
        \cite{lacher2015breast} & RGB-D & Microsoft Kinect & Kinect SDK + Custom & Any RGB-D camera$^*$ & \xmark & \xmark & \xmark \\
        \cite{henseler2016breast} & RGB-D & Microsoft Kinect & Kinect SDK + Custom & Any RGB-D camera$^*$ & \xmark & \xmark & \xmark \\
        \rowcolor{green!30}
        \cite{poehlmann2017breast} & RGB-D & Microsoft Kinect & Kinect SDK & Any RGB-D camera$^*$ & \cmark$^\dag$ & \cmark & \xmark \\
        \rowcolor{green!30}
        \cite{koban2018breast} & RGB-D & 3D scanner (\href{https://support.3dsystems.com/s/article/Sense-Scanner}{Sense}) & Supplied & Nothing & \cmark & \cmark & \xmark \\
        \cite{lacher2019breast} & RGB-D & Microsoft Kinect & Kinect SDK + Custom & Any RGB-D camera$^*$ & \xmark & \xmark & \xmark \\
        \rowcolor{green!30}
        \cite{oranges2019breast} & RGB-D & 3D scanner (\href{https://structure.io/structure-sensor-3}{Structure Sensor 3D}) & Supplied & Nothing & \cmark & \cmark & \xmark \\
        \cite{tong2020breast} & RGB-D & 2 mod. 3D scanners (HP Pro S3) & Supplied & Nothing & \xmark & \xmark & \xmark \\
        \cite{luu2021breast} & RGB-D & 3D scanner (Custom) & Custom & Nothing & \xmark & \xmark & \xmark \\
        \cite{pinto2022breast} & RGB-D & iPhone 11 (TrueDepth) & Scandy Pro & iPhone \& iPad w\textbackslash~TrueDepth & \xmark & \cmark & \xmark \\
        \cite{han2023breast} & RGB-D & iPhone 12 Pro (LiDAR) & Custom & iPhone \& iPad w\textbackslash~LiDAR$^\ddag$ & \cmark & \xmark & \xmark \\
        \cite{behrens2024breast} & RGB-D & iPhone 11 Pro Max (TrueDepth) & \href{https://3dscannerapp.com}{3D Scanner App} & iPhone \& iPad w\textbackslash~TrueDepth & \xmark & \cmark & \xmark \\
        \rowcolor{green!30}
        \cite{dijkman2024breast} & RGB-D & iPhone XR (TrueDepth) & \href{https://hege.sh}{Heges} & iPhone \& iPad w\textbackslash~TrueDepth & \cmark & \cmark & \xmark \\
        \cite{fu2024breast} & RGB-D & Intel RealSense & 3D Slicer + Custom & Any RGB-D camera$^*$ & \cmark & \xmark & \xmark \\
        \rowcolor{green!30}
        \cite{rudy2024breast} & RGB-D & iPhone X (TrueDepth) & Scandy Pro & iPhone \& iPad w\textbackslash~TrueDepth & \cmark & \cmark & \xmark \\
        \rowcolor{green!30}
        \cite{kyriazidis2025breast} & RGB-D & iPhone 15 Pro (LiDAR) & \href{https://3dscannerapp.com}{3D Scanner App} & iPhone \& iPad w\textbackslash~LiDAR & \cmark & \cmark & \xmark \\
        \midrule
        \cite{deHerasCiechomski2012breast} & Photo & Not specified & Custom & Any RGB camera & \cmark & \xmark & \xmark \\
        \cite{henseler2013breast} & Photo & 8 DSLR cameras & Dimensional Imaging 3D & Nothing & \xmark & \xmark & \xmark \\
        \cite{ruiz2018breast} & Photo & Not specified & Custom & Any RGB camera & \cmark & \xmark & \xmark \\
        \rowcolor{green!30}
        \cite{chrobot2025breast} & Photo & iPhone 15 & \href{https://3dscannerapp.com}{3D Scanner App} & Newer iPhone \& iPad & \cmark & \cmark & \xmark \\
        \midrule
        \rowcolor{green!30}
        Ours & Video & iPhone 12 Mini & Custom & Any RGB camera & \cmark & \cmark & \cmark \\
        \bottomrule
        \end{tabular}
    \end{adjustbox}
    \caption{\textbf{Overview of existing low-cost 3D breast surface reconstruction methods.} The list comprises all works that propose or use affordable alternatives to commercial 3D scanning systems from companies such as Artec and Canfield, all of which rely on \textit{commodity} and \textit{non-medical} hardware. Among these, our proposed method stands out as the only pipeline that is (i) compatible with the widest range of devices---that is, any device capable of recording a standard RGB video, (ii) handheld and publicly available without requiring the purchase of specialized hardware, and (iii) based entirely on open-source software. We highlight in green the most accessible pipelines, which we define as pipelines that are handheld (hence do not require extra equipment such as tripods) and rely only on publicly available software. $^*$Compatibility might require significant adaptation or exchange of the Kinect SDK; $^\dag$Multiple setups are proposed, and only some of them are handheld; $^\ddag$Compatibility unclear due to self-developed software (software was developed under iOS, hence compatibility is assumed for iPhone and iPad).}
    \label{tab:overview_breast_surface_reconstruction}
\end{table*}

\subsection{Breast Surface Reconstruction}
Next, we revisit 3D breast surface reconstruction pipelines, limiting our scope to low-cost approaches that utilize commodity hardware (\textit{i.e.}, methods proposing alternatives to commercial systems from Artec or Canfield).
A comprehensive overview of these works is provided in Table~\ref{tab:overview_breast_surface_reconstruction}.

\paragraph{RGB-D.}
The vast majority of the existing pipelines (19 out of 23) use RGB and depth data (referred to as \textit{RGB-D}), acquired using RGB-D cameras such as Microsoft Kinect \citep{costa2014breast,henseler2014breast,poehlmann2014breast,wheat2014breast,lacher2015breast,henseler2016breast,lacher2019breast} or Intel RealSense \citep{fu2024breast}.
In this context, \cite{lacher2017breast} presented a systematic comparison between Microsoft Kinect v1 and v2 and established 3D scanning solutions (Artec Eva and 3dMD\footnote{\url{https://www.3dmd.com}} stereophotogrammetry system) based solely on open-source software.
They found that both devices produce satisfactory 3D breast surface reconstructions within an error margin of 3 mm compared to the ground truth. 

\cite{lacher2019breast} introduced a \textit{non-rigid}, template-free method for surface reconstruction from an RGB-D stream captured with a Microsoft Kinect.
While being able to recover clinical-quality 3D breast scans without motion or breathing artifacts, long runtimes of 1--2 hours prohibit their method from widespread use in daily clinical practice. 

Recently, \cite{fu2024breast} proposed a portable setup involving the Intel RealSense D415 camera.
Although their method reportedly produces 3D breast scans with an average landmark error of less than 1.5 mm, they employ custom software not publicly available. 


Besides RGB-D cameras, researchers explored 3D breast surface reconstruction pipelines based on affordable (non-medical) structured light scanners, either commercially available and handheld \citep{koban2018breast,oranges2019breast} or custom-built and static \citep{tong2020breast,luu2021breast}.
All of these methods rely on proprietary software and require purchasing special hardware. 


A third recent line of work investigated the use of the iPhone's LiDAR sensor \citep{han2023breast,kyriazidis2025breast} and TrueDepth camera \citep{pinto2022breast,rudy2024breast,behrens2024breast,dijkman2024breast} for 3D breast surface reconstruction.
Both studies that utilize the iPhone's LiDAR sensor report very similar findings, evaluating surface reconstruction quality based on anthropometric distances measured between anatomical landmarks.
Most measurements show reasonable agreement with ground truth values obtained via measuring tape, with the notable exception of the nipple-to-inframammary fold distance.
This discrepancy is due to the low spatial resolution of the iPhone’s LiDAR sensor, which captures only coarse surface geometry and completely fails to represent high-frequency details---an issue clearly visible in the poor reconstructions shown in \cite{kyriazidis2025breast}.
On the other hand, the iPhone’s TrueDepth camera appears capable of reconstructing high-fidelity surfaces.
\cite{rudy2024breast} report an average surface-to-surface distance of approximately 1.5 mm between 3D breast reconstructions obtained with the TrueDepth camera and ground truth Vectra H2 scans.
These results confirm earlier findings by \cite{pinto2022breast}; however, \cite{dijkman2024breast} and \cite{behrens2024breast} report only sufficient reconstruction quality when assessed in the context of breast volume estimation.

In summary, among all low-cost depth-based pipelines, only the Microsoft Kinect, when paired with custom software \citep{lacher2019breast}, and the iPhone’s TrueDepth camera demonstrate the ability to reconstruct high-fidelity 3D breast surfaces, positioning them as a real alternative to commercial 3D scanning solutions.
Nevertheless, both approaches require special hardware and depend on either non-publicly available or proprietary software.

\paragraph{Photo.}
Only a few methods have been proposed for reconstructing 3D breast surfaces from multi-view images.
\cite{deHerasCiechomski2012breast} presented a web-based application for 3D breast surface reconstruction using three 2D photographs taken from frontal and lateral views, along with user-provided anthropometric measurements.
They reported a mean surface-to-surface reconstruction error ranging between 2 and 4 mm.
The method requires the user to select landmarks on \textit{all} three images (22 in total) and is not publicly available.

\cite{henseler2013breast} designed a static setup using eight DSLR cameras and employed photogrammetry to reconstruct the 3D surface of the breast.
Their setup is fairly large, complex, and impractical for everyday use, requiring significant effort and space to assemble.

Based on a custom-built 3D parametric breast model, \cite{ruiz2018breast} proposed a 3D surface reconstruction technique that operates on three 2D photographs captured from frontal and lateral perspectives ($\pm$ 90 degrees).
The method fits the parametric model to a sparse 3D point cloud reconstructed via triangulation of an automatically detected set of corresponding 2D landmarks across the three images.
Prior to model fitting, estimated camera parameters required for triangulation and the back-projected 3D landmarks are jointly refined using bundle adjustment.
From a technical perspective, their pipeline is very similar to our method, except for the fact that we obtain the sparse point cloud by applying Structure-from-Motion (SfM) to a set of multi-view images extracted from a video sequence instead of three images (note that they essentially also use SfM, but with a custom keypoint detector).
On the downside, they only evaluate their method based on the re-projection error of the 3D landmarks, leaving the actual quality of the reconstructed geometry unclear.
Furthermore, unlike ours, their pipeline is not publicly available. 

Recently, \cite{chrobot2025breast} employed an iPhone 15 and the 3D Scanner App's photogrammetry mode to reconstruct 3D breast surfaces from a series of 2D photographs, automatically taken every 0.8 seconds as controlled by the application.
They report only moderate accuracy, assessed using 14 anthropometric distances measured on the smartphone-based 3D reconstructions and ground truth Vectra H2 scans. 

In contrast to the aforementioned methods, our pipeline uses a simple monocular RGB video as input.
We chose to work with video sequences because they are easier and faster to acquire than individual photographs, thereby minimizing capture time and reducing motion and breathing artifacts.
Similar to \cite{ruiz2018breast}, we adopt a model-based 3D surface reconstruction approach, which is particularly robust in noisy settings, as expected due to the degraded image quality of video frames.
However, unlike their method, we employ a \textit{neural implicit} 3D breast model, enabling more detailed and accurate surface reconstructions.
Our approach is publicly available, relies solely on open-source software, and does not require specialized hardware---any standard RGB camera can be used.

\begin{figure}
    \centering
    \includegraphics[width=1\linewidth]{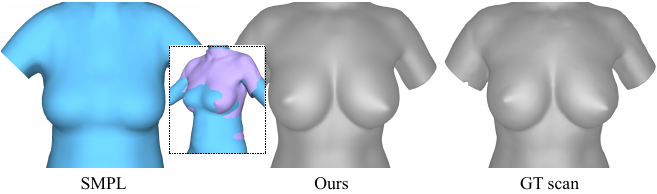}
    \caption{\textbf{Body model vs. breast-specific model}. We fit the female version of the popular SMPL model \citep{loper2015smpl} to one of our 3D breast scans, demonstrating that existing full-body models, despite including the chest region, can not accurately represent nude breast shapes.}
    \label{fig:smpl_vs_ours}
\end{figure}

\subsection{Human Body Models and Reconstruction}
We finally review related literature on human body models (that naturally include the breast region) and recent methods for neural implicit surface reconstruction from a sequence of multi-view RGB images.

\begin{figure*}[t]
	\centering
	\includegraphics[width=\linewidth]{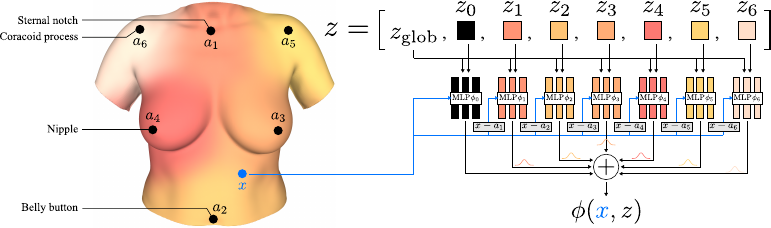}
	\caption{\textbf{Overview of our model's architecture.} Instead of using a single global MLP to represent breast shapes, following \cite{giebenhain2023nphm}, we partition the implicit domain into six smaller regions, centered around anchor points $\{a_1,a_2,\dots,a_6\}$ placed at anatomical landmark positions. Each region is represented by a shallow MLP $\phi_i$, conditioned on a local latent code $z_i$. To account for areas distant from any anchor, we introduce an additional MLP $\phi_0$ that shares the same representation as the local MLPs. The final SDF $\phi(x,z)$ at a query point $x$ is obtained by blending the outputs of all MLPs using a Gaussian weighting scheme. An illustration of the spatial influence of each local MLP is shown on the left, visualized on the resulting model's mean shape.}
    \label{fig:method_overview}
\end{figure*}

\paragraph{Models.}
Probably the most well-known full-body model is SMPL \citep{loper2015smpl}, a PCA-based parametric human pose and shape model that uses Linear Blend Skinning to allow for pose-dependent shape deformations and whose shape space is learned from the CAESAR \citep{robinette2002caesar} dataset. 
While the female version of SMPL does include the chest region, as demonstrated in Figure \ref{fig:smpl_vs_ours}, it is not able to accurately reconstruct nude breast shapes as it has been trained on 3D scans of women wearing a light bra. 
Numerous follow-ups improved upon SMPL, sticking to its mesh-based representation. 
Notably, GHUM \citep{xu2020ghum} replaces the PCA-based, linear shape space with a non-linear shape space learned using a Variational Auto Encoder.
Introduced by \cite{osman2020star}, STAR constrains SMPL's deformations to be more realistic, SUPR \citep{osman2022supr} uses a factorized representation based on part models, and SKEL \citep{keller2023skel} re-rigs SMPL with a biomechanics skeleton.

With the rise of neural implicit representations, Neural Parametric Models (NPMs; \cite{palafox2021npm}) propose a full-body model with disentangled human shape and pose, representing shape in canonical pose as the zero-level set of a latent-conditioned neural SDF, which is deformed through a learned deformation field to yield shapes in posed space.
NPMs differ from other neural implicit approaches \citep{deng2020nasa,mihajlovic2021leap,alldieck2021imghum,mihajlovic2022coap,palafox2022spam,mihajlovic2025volumetricsmpl} in that they do \textit{not} require any domain-specific annotations (such as a kinematic chain, skeleton, or part segmentations) or knowledge, therefore can be easily adapted to other parts of the human body.
Existing neural implicit body models are typically trained on human motion datasets captured from clothed subjects; hence, these models can not accurately represent breast shapes. 

Neural Parametric Head Models (NPHMs; \cite{giebenhain2023nphm}) built upon NPMs and proposed a part-based head model, decomposing the implicit head domain into multiple smaller regions each of which is represented by a \textit{local} neural SDF instead of a single global network.
Facial expressions are modeled as deformations of shape in canonical space, analogously to pose in NPMs.
Conceptually similar to \citep{zheng2022imface}, this strategy significantly increases the level of detail, reaching state-of-the-art results in dynamic head reconstruction from monocular RGB videos \citep{giebenhain2024mononphm}.

In this work, we adopt NPHMs' architecture and use it to model breast shapes.
We note that, in this context, the previously introduced iRBSM \citep{weiherer2025irbsm} follows the architecture of NPMs without pose deformations, whereas our proposed model, liRBSM, resembles NPHMs' architecture but omits the facial expression network. 

\paragraph{Reconstruction.}
Most of the state-of-the-art neural surface reconstruction methods employ a \textit{prior-free} approach, combining implicit surface representations (such as occupancy functions or SDFs) with differentiable volume rendering \citep{niemeyer2020dvr,lior2021volsdf,wang2021neus,oechsle2021unisurf,zehao2022monosdf,zhaoshuo2023neuralangelo,wang2023neus2,xu2024supernormal}.
Per-scene optimization methods typically suffer from long run-times and, due to missing geometric priors, degenerate quickly if only a sparse set of images is available.

In contrast, \textit{prior-based} approaches work well in a few-shot scenario.
More recent methods combine differentiable volume rendering with neural parametric models \citep{ramon2021h3d,giebenhain2024mononphm} and optional displacement fields for improved geometrical detail \citep{grassal2022nha,caselles2025sirapp} and are primarily being developed in the head domain.
Fitting a model to a set of images is then done in an analysis-by-synthesis fashion, and requires a careful initialization, (facial) landmarks and/or silhouette masks detected in \textit{every} image, and photometric losses to supervise the optimization process.

Contrary to rendering-based 3D reconstruction methods and inspired by the recent success of differentiable SfM pipelines \citep{wang2024duster,wang2024vggsfm,duisterhof2025mastersfm,wang2025vggt} in terms of robustness against photometric changes and few-shot capabilities, we opted for a more classical approach and fit our 3D neural parametric breast model directly onto a \textit{point cloud} obtained from SfM.
This model-based surface reconstruction strategy is robust against noise but retains the few-shot capabilities of prior-based approaches while avoiding cumbersome per-image landmark detection, which is difficult in the breast domain due to the lack of automatic landmark detectors.
Moreover, it bypasses the necessity of large paired 2D-3D datasets common in human mesh recovery \citep{tian2023hmr}, as no such data is available in the breast domain.
As we will show later, our method achieves state-of-the-art results and works well with as few as three images extracted from the input video.

\section{Methods}
We will now describe our model and the proposed 3D surface reconstruction pipeline in detail.

\subsection{Localized Neural Implicit Breast Model}
Our model adopts the neural implicit representation of the iRBSM, which represents breast shapes as the zero-level set of a latent-conditioned neural SDF.
However, inspired by recent state-of-the-art face models \citep{zheng2022imface,giebenhain2023nphm,potamias2025imhead} and to increase the level of detail, we further employ a space partitioning approach and decompose the implicit breast domain into multiple smaller regions, each represented by a \textit{local} SDF anchored at anatomical landmark positions. 
Specifically, as illustrated in Figure \ref{fig:method_overview}, instead of a single global MLP that models the entire breast geometry, we use an ensemble  of $K$ \textit{local} MLPs
\begin{equation}
    \begin{split}
    \phi_k:\mathbb{R}^3\times\mathbb{R}^{L_\text{glob}}\times\mathbb{R}^{L_\text{loc}}&\rightarrow\mathbb{R}\\
        (x,z_\text{glob},z_k)&\mapsto\phi_k(x-a_k,z_\text{glob},z_k)
    \end{split}
    \label{eq:local_mlps}
\end{equation}
centered around anchor points $a_k\in\mathbb{R}^3$ and blended to yield a parametric model of the form
\begin{equation}
    \phi(x,z)=\sum_{k=1}^Kw_k(x)\phi_k(x-a_k,z_\text{glob},z_k),
    \label{eq:model_formulation}
\end{equation}
where $z_\text{glob}\in\mathbb{R}^{L_\text{glob}}$ is a global (per-shape) latent code, $z_k\in\mathbb{R}^{L_\text{loc}}$ are local latent codes for the individual MLPs, and $z:=[z_\text{glob},z_1,z_2,\dots,z_K]$.
The blending weights $w_k(x)$ in Eq. (\ref{eq:model_formulation}) are defined as
\begin{equation}
    w_k(x)=\frac{w^*_k(x)}{\sum_{k'=1}^Kw^*_{k'}(x)}, 
\end{equation}
where
\begin{equation}
    w^*_k(x)=\exp\left(-\frac{\Vert x-a_k\Vert^2_2}{2h^2}\right)
    \label{eq:blending}
\end{equation}
is an isotropic Gaussian kernel with bandwidth $h>0$.
Anchor points $a=[a_1,a_2,\dots,a_K]$ are predicted using a small MLP $\phi_\text{anc}:\mathbb{R}^{L_\text{glob}}\rightarrow\mathbb{R}^{K\times 3}$ conditioned on the global latent code, \textit{i.e.}, $a=\phi_\text{anc}(z_\text{glob})$.

Finally, since each of the local MLPs focuses on a specific part of the breast, following \cite{giebenhain2023nphm}, we add an additional MLP to our formulation in Eq. (\ref{eq:model_formulation}) that captures global context far away from the anchor points (effectively resulting in $K+1$ regions in which we divide the domain). 
This MLP shares the same representation as the local MLPs introduced in Eq. (\ref{eq:local_mlps}), is anchored at the origin, and uses a constant weight response. 
We refer to this network as \textit{background MLP}.

\begin{figure*}[t]
	\centering
	\includegraphics[width=\linewidth]{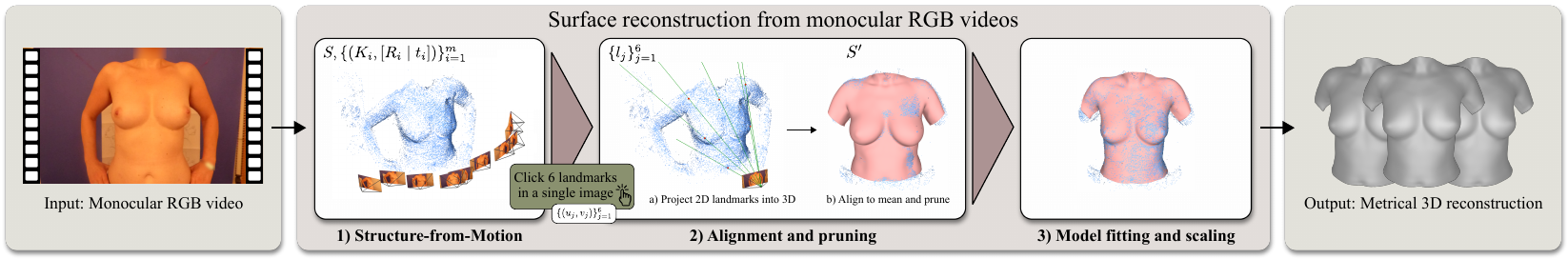}
	\caption{\textbf{Surface reconstruction from monocular RGB videos.} We present a method for low-cost \textit{and} accurate metrical breast surface reconstruction from just a single monocular RGB video, acquired using commodity hardware (such as smartphones). Our pipeline starts by applying Structure-from-Motion to a set of frames extracted from the input video, resulting in a sparse point cloud and camera parameters. We then align the estimated point cloud to our model's mean shape, prune away background points, fit our model to the resulting point cloud, and finally recover real-world scale. Depending on the available hardware, the entire reconstruction can be executed in under six minutes.}
    \label{fig:sfm_pipeline}
\end{figure*}

\subsubsection{Training}
We train our model in an auto-decoder fashion \citep{park2019deepsdf}, jointly optimizing the network's parameters and latent codes. 
Given a dataset of 3D breast scans, each of which is represented as normalized and centered point cloud $X=\{x_1,x_2,\dots,x_n\}\subset[-1,1]^3$ with normals $N=\{n_1,n_2,\dots,n_n\}\subset\mathbb{R}^3$ and ground truth anchor points $a_\text{gt}\in\mathbb{R}^{K\times 3}$ placed at expert-annotated anatomical landmark positions, we minimize the same loss function as in \citep{giebenhain2023nphm}, namely,
\begin{equation}
    \mathcal{L}=\mathcal{L}_\text{SDF}+\lambda_5\mathcal{L}_\text{anc}+\lambda_6\Vert z\Vert_2,
\end{equation}
where
\begin{multline}
    \mathcal{L}_\text{SDF}=\sum_{i=1}^n\lambda_1|\phi(x_i,z)|+\lambda_2\Vert\nabla_{x_i}\phi(x_i,z)-n_i\Vert_2+\\
    \mathbb{E}_{x\sim D}\left[\lambda_3|\Vert\nabla_x\phi(x,z)\Vert_2-1|+\lambda_4\exp(-\alpha|\phi(x,z)|)\right]
    \label{eq:sdf_loss}
\end{multline}
ensures that $\phi$ represents a valid SDF,
\begin{equation}
    \mathcal{L}_\text{anc}=\Vert\phi_\text{anc}(z_\text{glob})-a_\text{gt}\Vert_2
    \label{eq:anchor_loss}
\end{equation}
is used to supervise anchor predictions, and the last term regularizes latent codes.
The first term in Eq. (\ref{eq:sdf_loss}) enforces that the predicted SDF is zero at the given surface points and that its gradients on the zero-level set match the ground truth surface normals.
The second term serves as a regularizer, consisting of an Eikonal constraint that encourages the SDF’s gradient to have unit norm almost everywhere, and a volume loss (with parameter $\alpha\gg1$) that penalizes off-surface points for creating SDF values close to zero.
Off-surface points $D\subset\mathbb{R}^3$ are obtained by randomly offsetting the surface points and adding some points uniformly sampled in $D:=[-1,1]^3\cup\{x_1+\epsilon_1,x_2+\epsilon_2,\dots,x_n+\epsilon_n\}$ with $\epsilon_i\sim\mathcal{N}(0,\sigma^2)$.
We refer the reader to \citep{gropp2020geomInit} and \citep{sitzmann2020siren} for further details about this loss.

\subsubsection{Inference}
\label{subsubsec:inference}
Given a 3D breast scan as \textit{unoriented} point cloud (without normals) $X'=\{x'_1,x'_2,\dots,x'_{n'}\}$ uniformly scaled to $[-1,1]^3$ and centered at the origin, at test-time, we obtain the corresponding latent code $z^*$ via maximum a posteriori estimation as proposed in \citep{park2019deepsdf}.
In particular, we fix the model's parameters and optimize
\begin{equation}
    z^*=\argmin_{z\in\mathbb{R}^L}\left\{\sum_{i=1}^{n'}|\phi(x'_i,z)|+\lambda\Vert z\Vert_2\right\},
    \label{eq:latent_optim}
\end{equation}
where $L=L_\text{glob}+(K+1)L_\text{loc}$ denotes the total latent dimensionality of the model, and $\lambda\geq0$ can be used to filter out noise. 
We then apply the inverse scale and translation to the extracted mesh in order to recover its original position and scale.

\subsection{Surface Reconstruction From Monocular RGB Videos}
Given a sequence of RGB images $I=\{I_1,I_2,\dots,I_m\}$ extracted from the input video, along with a set of 2D landmarks clicked in a \textit{single} frame, our goal is to 3D reconstruct a metrically correct surface mesh, \textit{i.e.}, infer the model's parameters $z$ that best explains the underlying surface. 
We approach this problem by first applying Structure-from-Motion (SfM) to $I$, and then robustly fitting our model to the resulting (potentially noisy and) arbitrarily scaled sparse point cloud by solving Eq. (\ref{eq:latent_optim}).

In particular, our pipeline, as shown in Figure \ref{fig:sfm_pipeline}, is as follows.
We begin by extracting a set $I$ of $m$ RGB images from the input video.
To ensure that the selected frames are both sharp and evenly distributed across the video's timeline, we adopt a window-based selection strategy.
Specifically, we first select $m$ temporally equidistant candidate frames (regardless of image quality) and then search within a local neighborhood around each candidate for a frame that ranks within the top 25\% of the overall sharpest frames.
If no such frame is found, we adaptively relax the sharpness threshold over three iterations before falling back to simply selecting the sharpest frame within the window.
This approach ensures good temporal coverage while maintaining consistent image sharpness within \textit{and} across local windows.

Next, we use a state-of-the-art and off-the-shelf SfM pipeline, VGGSfM \citep{wang2024vggsfm}, to reconstruct camera parameters and a sparse point cloud $S\subset\mathbb{R}^3$ from the given images, $I$.

After that, we align the point cloud to the model's mean shape using a set of $K$ 2D landmarks annotated in just a \textit{single} image. 
Let this set of landmarks be denoted as $\{(u_1,v_1),(u_2,v_2),\dots,(u_K,v_K)\}\subset\mathbb{N}^2$, and let us further assume that these landmarks have been annotated in the $i$-th image of $I$.
We obtain corresponding 3D landmark positions $l_j$ by back-projecting $(u_j,v_j)$ based on the estimated camera parameters. 
To do so, we cast rays $r_j(t)=o+td_j$ from the camera's origin $o\in\mathbb{R}^3$ through pixel $(u_j,v_j)$ into the scene, selecting the point in $S$ as back-projected landmark that first hits the point cloud.
More formally, we choose the point in $S$ that is closest to the camera and near the ray, setting
\begin{equation}
    l_j=\argmin_{x\in S}\left\{t^j_x=(x-o)\cdot d_j: d(x,r_j)\leq\delta\right\}
    \label{eq:back-projection}
\end{equation}
for all $j\in\{1,2,\dots,K\}$, where $t^j_x$ denotes the distance from $x$ to the camera along the ray $r_j$ (\textit{i.e.}, the depth),
\begin{equation}
    d(x,r_j)=\Vert(x-o)\times d_j\Vert_2
\end{equation}
is the distance from point $x$ to ray $r_j$, and $\delta>0$.
Moreover,
\begin{equation}
    o=-R_i^\top t_i,\quad d_j=\frac{\hat{d}_j}{\Vert\hat{d}_j\Vert_2},\quad\hat{d}_j=R_i^\top K_i^{-1} 
    \begin{bmatrix}
    u_j \\
    v_j \\
    1
    \end{bmatrix}.
\end{equation}
Here, $K_i\in\mathbb{R}^{3\times3}$ denotes the estimated camera intrinsics of the $i$-th image, and $[R_i\mid t_i]\in\mathbb{R}^{3\times 4}$ are the extrinsic parameters, formed by rotation $R_i\in\text{SO}(3)$ and translation $t_i\in\mathbb{R}^3$.
Finally, we globally align the estimated point cloud $S$ to the model's mean shape by computing a similarity transformation between the back-projected 3D landmarks $\{l_1,l_2,\dots,l_K\}$ and the corresponding model's average anchor points using the method in \citep{umeyama1991procrustes}.
Once aligned, we prune away points in $S$ that are further away from the mean shape than a pre-selected threshold, $\tau>0$:
\begin{equation}
    S'=\{x\in S:\Vert x-c_M(x)\Vert_2\leq\tau\},
    \label{eq:pruning}
\end{equation}
where $c_M(x)=\argmin_{x'\in M}\{\Vert x-x'\Vert_2\}$ is the closest point to $x$ in the mean shape, $M\subset\mathbb{R}^3$.
This effectively removes unwanted points in the background. 

As a last step, we fit our model to the aligned and pruned point cloud $S'$ as detailed in Section \ref{subsubsec:inference} (without the additional scaling and translation step), resulting in a 3D surface mesh that lives in $[-1,1]^3$.
For a lot of applications, including anthropometry and breast volume estimation, however, a metrical (\textit{i.e.}, real-world scale) reconstruction is required.
To recover the correct scale, we propose to use either of the following two strategies:
\begin{enumerate}
    \item Scale the reconstructed mesh based on a known landmark distance (we use the nipple-to-nipple distance) measured on the real subject. This yields an \textit{exact metrical reconstruction}, but requires additional data collection. 
    \item Scale the reconstructed mesh with the inverse average scaling factor obtained from our real-world scale training data when normalized to $[-1,1]^3$. This approach does \textit{not} yield exact metrical reconstructions on a per-instance level---it merely results in a statistically \textit{approximate metrical reconstruction}; at the same time, however, it avoids any additional data collection. 
\end{enumerate}

As we will show in our experimental evaluation, both strategies result in nearly the same reconstruction accuracy, indicating that approximate metrical reconstructions are of sufficient quality in practice.

\subsection{Implementation Details}
\label{subsec:implementation_details}

\paragraph{Model.}
We use $K=6$ anchor points, distributed as shown in Figure \ref{fig:method_overview}.
Furthermore, our model employs a global latent dimension of $L_\text{glob}=128$, and a local latent dimension of $L_\text{loc}=64$ (as such, our model has a total latent dimension of $L=128+(6+1)64=576$).
The bandwidth $h$ of the Gaussian kernel in Eq. (\ref{eq:blending}) is set to 0.25, and a constant weight response of 0.2 is used for the background MLP.
The remaining model and training parameters closely resemble those from \citep{giebenhain2023nphm}.
In particular, local MLPs $\phi_k$ are fully connected and have four 200-dimensional hidden layers with a skip connection to the middle layer. 
We use the geometric initialization scheme proposed in \citep{gropp2020geomInit}.
The final SDF value is regressed by applying the softplus activation function. 
The MLP that predicts anchor positions has a single, 256-dimensional hidden layer with ReLU activation. 
Latent codes are initialized from a zero-mean normal distribution with variance $10^{-4}$. 
We trained until convergence, but not longer than 15,000 epochs, using a batch size of 16 and the AdamW optimizer \citep{loshchilov2019adamw} with a weight decay of 0.01 and a learning rate of $5\times 10^{-4}$ for the model parameters and $10^{-3}$ for latent codes. 
Both learning rates are decayed by a factor of 0.5 every 3,000 epochs. 
We use 500 on-surface points and 500 off-surface points sampled from $D$ (obtained using $\sigma^2=0.01$) for each shape.
We empirically set $\lambda_1=2$, $\lambda_2=0.3$, $\lambda_3=0.1$, $\lambda_4=0.01$, $\lambda_5=7.5$, $\lambda_6=0.01$, and $\alpha=10$.
Gradients are clipped with a cut-off value of 0.1.
Training our model took about 18 hours on a single NVIDIA A40 with 40 GB of VRAM.

At inference, we optimize Eq. (\ref{eq:latent_optim}) for 1,000 iterations using Adam \citep{kingma2015adam} with a learning rate of $10^{-2}$.
Additionally, we decay the learning rate by a factor of 0.5 every 200 iterations and divide $\lambda$ by 3 after 200 iterations and by 10 after 600 iterations.
Inference takes about 15 seconds, measured on a single NVIDIA RTX A4000 with 20 GB of VRAM and a resolution of $256^3$.

\paragraph{Surface Reconstruction.}
We assume shared camera intrinsics across all images during SfM, which is safe to accept since our images are extracted from a video sequence. 
Moreover, we slightly adapted VGGSfM's default parameters (see discussion below).
We selected six landmarks in a single image, corresponding to the model's anchor points placed at the anatomical locations and described below.
Furthermore, $\delta$ is set such that the constraint in Eq. (\ref{eq:back-projection}) considers the five closest points to a ray.
Lastly, $\tau=0.2$ in Eq. (\ref{eq:pruning}), effectively discarding points that are further away from the mean shape than $\sim$10 cm.

\section{Experiments and Results}
We performed extensive experiments to validate our new model and the proposed 3D surface reconstruction pipeline.

\paragraph{Data.}
Our model is trained on the pre-processed dataset of the iRBSM \citep{weiherer2025irbsm}, which includes 168 consistently oriented 3D breast scans.
All scans have been taken in a standing position using Canfield's Vectra H2 system. 
Subsequently, after data acquisition, the following six anatomical landmarks have been expert-annotated in each 3D breast scan: (1) sternal notch, (2) belly button, (3) left nipple, (4) right nipple, (5) left coracoid process, and (6) right coracoid process, see also Figure \ref{fig:method_overview}.

To evaluate our model and the proposed surface reconstruction pipeline, we collected a test dataset of ten 3D breast scans acquired in a standing position using the Vectra H2 system, along with corresponding monocular RGB video sequences.
The videos were captured using an iPhone 12 Mini while moving in a 180-degree circular arc around the subject at an approximately constant speed.
No specific constraints were imposed during the data acquisition process, except that the region between the belly button and the sternal notch (both \textit{included}) remained visible throughout the capture.
On average, recording the videos took about 20 seconds per subject.

\paragraph{Baselines.} We compare our model against the publicly available RBSM \citep{weiherer2023rbsm} and the recently proposed iRBSM \citep{weiherer2025irbsm}. The mesh-based RBSM is fitted to point clouds $X'$ using model-based non-rigid surface registration, optimizing
\begin{equation}
    L(\alpha)=\Vert x(\alpha)-c_{x(\alpha)}(X')\Vert_2+\lambda_1\Vert x_J(\alpha)-l\Vert_2+\lambda_2\Vert\alpha\Vert_2
    \label{eq:rbsm_fitting}
\end{equation}
for 2,000 iterations using Adam \citep{kingma2015adam}, a learning rate of $10^{-2}$, $\lambda_1=0.1$, and $\lambda_2=0.01$.
The learning rate is decayed by a factor of 0.5 every 500 iterations.
In Eq. (\ref{eq:rbsm_fitting}), $x(\alpha)=\bar{x}+U\alpha$ denotes the classical PCA-based statistical shape model, where $\bar{x}\in\mathbb{R}^{3n}$ is the mean shape and $U\in\mathbb{R}^{3n\times q}$ holds the model's $q$ principal components.
Furthermore, $J\subset\mathbb{N}$ is an index set, selecting the $m$ landmarks in the model that correspond to $l\in\mathbb{R}^{3m}$ on the target, $X'$.
Due to the absence of the belly button and coracoid process in the RBSM, we use the following five landmarks to guide the registration process: sternal notch, left nipple, right nipple, left lower breast pole, and right lower breast pole; see, \textit{e.g.}, \citep{hartmann2020breast} for a detailed explanation of these landmarks. 
We employ the augmented version of the RBSM that also includes mirrored scans, and use all $q=219$ available principal components.

The iRBSM is fitted by solving Eq. (\ref{eq:latent_optim}) under the same setup used for our model, as detailed in Section \ref{subsec:implementation_details}.

\paragraph{Metrics.}
We report Chamfer distance (CD), F-Score with a threshold of 2.5 mm, and normal consistency (NC) to compare our surface reconstructions against ground truth 3D breast scans (we refer to Appendix \ref{app:metrics} for an exact definition of these metrics).
To ensure fair comparison, we compute metrics only within the breast region, defined by cropping meshes using axis-aligned planes passing through the upper breast poles, along axillary lines, and approximately 5 cm below the lower breast poles.

\begin{table}[t]
    \begin{tabular*}{\linewidth}{l@{\extracolsep\fill}ccc}
    \toprule
    & CD $\downarrow$ & F-Score $\uparrow$ & NC $\uparrow$ \\
    \midrule
    RBSM & 3.40 {\tiny $\pm$ 1.20} & 52.0 {\tiny $\pm$ 19.67} & 96.8 {\tiny $\pm$ 1.73} \\
    iRBSM & 1.13 {\tiny $\pm$ 0.38} & 93.5 {\tiny $\pm$ 5.55} & 99.0 {\tiny $\pm$ 0.96} \\
    Ours & \textbf{0.77} {\tiny $\pm$ 0.19} & \textbf{98.6} {\tiny $\pm$ 2.43} & \textbf{99.6} {\tiny $\pm$ 0.47} \\
    \bottomrule
    \end{tabular*}
    \caption{\textbf{Intrinsic model evaluation}. We show quantitative results for surface reconstruction from clean point clouds consisting of 5,000 points. Our model outperforms the RBSM and iRBSM by a large margin.}
    \label{tab:results_clean}
\end{table}

\begin{figure*}[t]
	\centering
	\includegraphics[width=\linewidth]{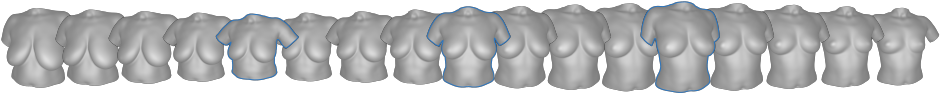}
	\caption{\textbf{Latent space interpolation.} We linearly interpolate between the latent codes corresponding to the shapes outlined in blue. Starting from the mean shape at the center, we interpolate in positive direction along the first and second principal components on the right, and in the negative direction on the left. Principal components are derived by applying PCA to the optimized latent codes from the training data. Our latent space is continuous and well-behaved.}
    \label{fig:latent_code_interpolation}
\end{figure*}

\begin{figure}
    \centering
    \includegraphics[width=\linewidth]{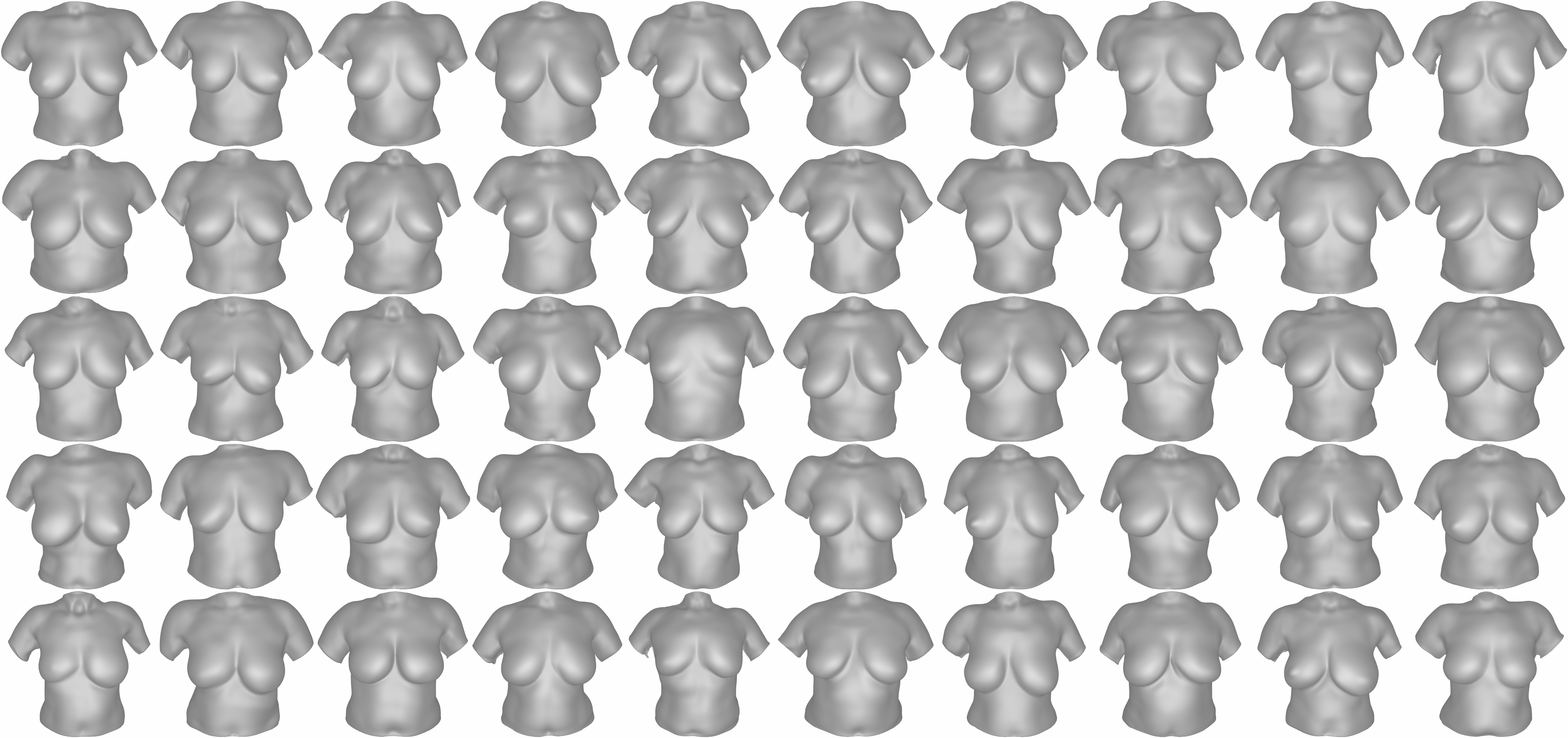}
    \caption{\textbf{Random samples from our model.} The proposed model is highly expressive and allows for a variety of plausible breast shapes.}
    \label{fig:random_samples}
\end{figure}

\subsection{Intrinsic Model Evaluation}
\label{subsec:intrisic_evaluation}
We begin by evaluating our model’s ability to reconstruct \textit{clean} point clouds, which are obtained through random sampling of 5,000 surface points from each of our test scans.
This experiment is designed to measure the true representational capacity of our model in comparison to the RBSM and iRBSM, as surface reconstruction quality is assessed independently of any compromised (noisy) or incomplete data; hence, we refer to it as \textit{intrinsic model evaluation}.
We set $\lambda=0.05$.

Results are summarized in Table \ref{tab:results_clean}.
Our model consistently outperforms both baselines across all evaluation metrics.
Compared to the mesh-based RBSM, it achieves over a four times reduction in CD, indicating a significantly more accurate reconstruction of the underlying surface geometry.
Furthermore, our localized model surpasses the global iRBSM by reducing CD by more than 30\%, demonstrating the effectiveness of incorporating spatial locality.
We also investigated the impact of the number of input points during surface fitting and found that both the iRBSM and our model benefit from increased point density. 
Please see Appendix \ref{app:intrinsic_model_additional} for quantitative and qualitative results.

Finally, we study our model's shape space by inspecting random samples and linear interpolants in latent space. 
As seen from Figures \ref{fig:latent_code_interpolation} and \ref{fig:random_samples}, our model is highly expressive and its shape space is continuous and well-behaved, being able to generate a variety of plausible and realistic-looking breast shapes.
 
\begin{table}
    \begin{adjustbox}{width=\linewidth}
        \begin{tabular}{lccc}
        \toprule
        & CD $\downarrow$ & F-Score $\uparrow$ & NC $\uparrow$ \\
        \midrule
        iRBSM ($-\mid-$; 256) & 1.13 {\tiny $\pm$ 0.38} & 93.5 {\tiny $\pm$ 5.55} & 99.0 {\tiny $\pm$ 0.96} \\
        $K=4$ ($96\mid 32$; 256) & 0.93 {\tiny $\pm$ 0.24} & 96.6 {\tiny $\pm$ 3.69} & 99.4 {\tiny $\pm$ 0.59} \\
        $K=6$ ($32\mid 32$; 256) & 0.95 {\tiny $\pm$ 0.28} & 96.3 {\tiny $\pm$ 4.31} & 99.3 {\tiny $\pm$ 0.73} \\
        $K=6$ ($64\mid 32$; 288) & 0.89 {\tiny $\pm$ 0.24} & 97.5 {\tiny $\pm$ 3.67} & 99.4 {\tiny $\pm$ 0.70} \\
        $K=6$ ($96\mid 32$; 320) & 0.87 {\tiny $\pm$ 0.21} & 97.7 {\tiny $\pm$ 2.88} & 99.5 {\tiny $\pm$ 0.54} \\
        \midrule
        \shortstack[l]{\scriptsize Ours \\ $K=6$ ($128\mid 64$; 576)} & \textbf{0.77} {\tiny $\pm$ 0.19} & \textbf{98.6} {\tiny $\pm$ 2.43} & \textbf{99.6} {\tiny $\pm$ 0.47} \\
        \bottomrule
        \end{tabular}
    \end{adjustbox}
    \caption{\textbf{Ablation on anchor layout and latent dimensions.} Numbers in parentheses represent global, local, and total latent dimension, \textit{i.e.}, $(L_\text{glob}\mid L_\text{loc};L)$. Proposed localized models consistently outperform the global iRBSM.}
    \label{tab:ablation}
\end{table}

\subsection{Ablations}
\label{subsec:ablations}
Next, we ablate key design choices, the number of anchor points, and the model's global and local latent dimensions.

Quantitative results can be found in Table \ref{tab:ablation}, and we refer to Appendix \ref{app:ablation_additional} for qualitative comparisons.
Interestingly, when the total latent dimensionality of our models is matched to that of the iRBSM ($L = 256$), the model with four anchors slightly outperforms the six-anchor setup (0.93 vs. 0.95 in CD).
However, when both the global \textit{and} local latent dimensionalities are matched across configurations ($L_{\text{loc}}=96$, $L_{\text{glob}}=32$) and the total latent dimension is disregarded, the six-anchor model significantly outperforms the four-anchor variant (0.87 vs. 0.93).
In general, as the global latent dimensionality in the six-anchor configuration increases, the model begins to surpass the four-anchor variant, with reconstruction quality improving steadily with larger $L_{\text{glob}}$.
Finally, all of our models consistently outperform the global iRBSM, regardless of anchor configuration or latent dimensionality.

\begin{table}
    \begin{tabular*}{\linewidth}{l@{\extracolsep\fill}ccc}
    \toprule
    & CD $\downarrow$ & F-Score $\uparrow$ & NC $\uparrow$ \\
    \midrule
    \multicolumn{4}{l}{Non-metrical evaluation} \\
    \midrule
    \hspace{0.3mm} iRBSM & 2.06 {\tiny $\pm$ 1.01} & 78.1 {\tiny $\pm$ 16.97} & 98.6 {\tiny $\pm$ 1.46} \\
    \hspace{0.3mm} Ours w\textbackslash o $\mathcal{L}_\text{anc}$ & 1.74 {\tiny $\pm$ 0.65} & 86.0 {\tiny $\pm$ 11.17} & 99.2 {\tiny $\pm$ 0.71} \\
    \hspace{0.3mm} Ours & \textbf{1.73} {\tiny $\pm$ 0.65} & \textbf{86.1} {\tiny $\pm$ 11.15} & \textbf{99.2} {\tiny $\pm$ 0.70} \\
    \midrule
    \multicolumn{4}{l}{Metrical evaluation} \\
    \midrule
    \multicolumn{4}{l}{\small----- Exact metrical -----------------------------------------------------} \\
    \hspace{0.3mm} iRBSM & 2.28 {\tiny $\pm$ 0.65} & 70.1 {\tiny $\pm$ 12.36} & 98.5 {\tiny $\pm$ 1.34} \\
    \hspace{0.3mm} Ours w\textbackslash o $\mathcal{L}_\text{anc}$ & 1.97 {\tiny $\pm$ 0.51} & 76.5 {\tiny $\pm$ 10.86} & 99.0 {\tiny $\pm$ 0.75} \\
    \hspace{0.3mm} Ours & \textbf{1.96} {\tiny $\pm$ 0.47} & \textbf{76.7} {\tiny $\pm$ 10.05} & \textbf{99.0} {\tiny $\pm$ 0.68} \\
    \multicolumn{4}{l}{\small----- Approximate metrical --------------------------------------------} \\
    \hspace{0.3mm} iRBSM & 2.30 {\tiny $\pm$ 0.77} & 70.0 {\tiny $\pm$ 13.60} & 98.5 {\tiny $\pm$ 1.34} \\
    \hspace{0.3mm} Ours w\textbackslash o $\mathcal{L}_\text{anc}$ & 1.98 {\tiny $\pm$ 0.47} & 76.2 {\tiny $\pm$ 10.84} & 99.0 {\tiny $\pm$ 0.75} \\
    \hspace{0.3mm} Ours & \textbf{1.97} {\tiny $\pm$ 0.47} & \textbf{76.4} {\tiny $\pm$ 10.00} & \textbf{99.0} {\tiny $\pm$ 0.68} \\
    \bottomrule
    \end{tabular*}
    \caption{\textbf{Quantitative results for 3D surface reconstruction from monocular RGB videos.} The proposed pipeline in combination with our new model performs best.}
    \label{tab:results_video}
\end{table}

\begin{figure*}[t]
    \centering
    \includegraphics[width=\linewidth]{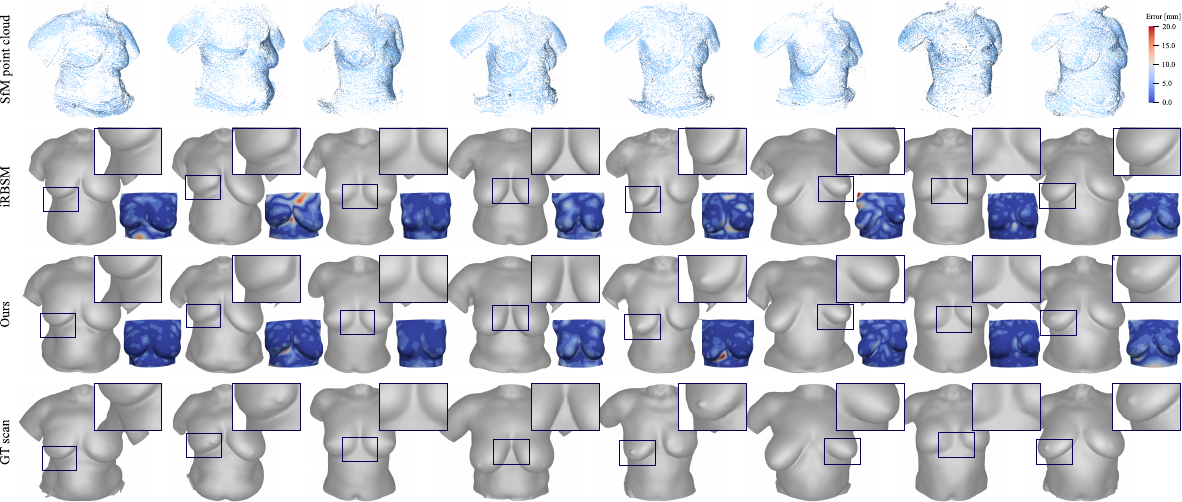}
    \caption{\textbf{Qualitative results for surface reconstruction from monocular RGB videos.} We show surface reconstructions obtained from fitting the iRBSM and our model to point clouds acquired by applying SfM to monocular RGB videos. In contrast to the iRBSM, our model is capable of recovering fine anatomical details such as skin folds (1st column) and nipples (5th and 8th columns), and is generally better at inferring the correct underlying shape, as seen in the remaining columns.}
    \label{fig:qual_results_video}
\end{figure*}

\subsection{Surface Reconstruction From Monocular RGB Videos}
Finally, we evaluate our proposed 3D surface reconstruction pipeline.
Since the reconstructed surfaces and ground truth 3D breast scans are not precisely aligned in this setting---unlike in the clean point cloud fitting scenario described in Section \ref{subsec:intrisic_evaluation}, where points are sampled directly from the ground truth surface---we follow common practice \citep{sanyal2019now} and apply landmark-based Procrustes alignment to align the reconstructed surface to the ground truth 3D breast scan, followed by an iterative closest point (ICP) algorithm.
Similar to \citep{zielonka2022metrical}, we evaluate performance under two settings, non-metrical and metrical.
In the non-metrical setting, we disregard the fact that our method produces metrically correct 3D reconstructions and allow the Procrustes and ICP steps to estimate a global scaling factor in addition to rotation and translation. 
In the metrical setting, we respect the real-world scale of the method's output, restricting alignment to be purely rigid (only rotation and translation).
Once aligned, evaluation metrics are computed as described above.
Metrics obtained from the non-metrical evaluation measure the true discrepancy between reconstructed and ground truth geometry, whereas numbers obtained from the metrical evaluation better reflect the overall reconstruction quality when results are to be used within a real-world application that requires a metrical context (such as breast volume estimation, for example).
We use $\lambda=0.1$ during surface fitting and further employ the anchor loss in Eq. (\ref{eq:anchor_loss}), weighted with 0.1, as we found that this slightly improves our reconstructions.

Results are shown in Table \ref{tab:results_video}, obtained by extracting 30 frames from the input video over time.
The introduced pipeline in combination with the proposed model is able to recover accurate 3D breast surfaces, reaching a CD of less than 2 mm in the metrical setting compared to 2.3 mm when using the iRBSM for surface reconstruction.
This improvement is also evident visually, as shown in Figure \ref{fig:qual_results_video}.
Our model yields high-fidelity 3D surface reconstructions, successfully capturing even fine anatomical details such as skin folds and nipples.
Moreover, the point clouds obtained via SfM appear remarkably accurate, despite the relatively low quality of the extracted video frames, which suffer from noise and suboptimal lighting during capture.
We attribute this robustness to VGGSfM’s learning-based approach, as we found that traditional methods such as COLMAP \citep{schoenberger2016sfm} produce \textit{significantly} noisier and sparser point clouds, even after moderate parameter tuning.
Interestingly, when using our first strategy to obtain real-world scale reconstructions (which requires the measurement of an additional landmark distance on the real subject), CD decreases only by 0.01 mm, suggesting that approximate metrical reconstructions are sufficient in practice. 

\subsubsection{Ablations}
\label{subsubsec:pipeline_ablations}

\begin{figure}
    \centering
    \includegraphics[width=\linewidth]{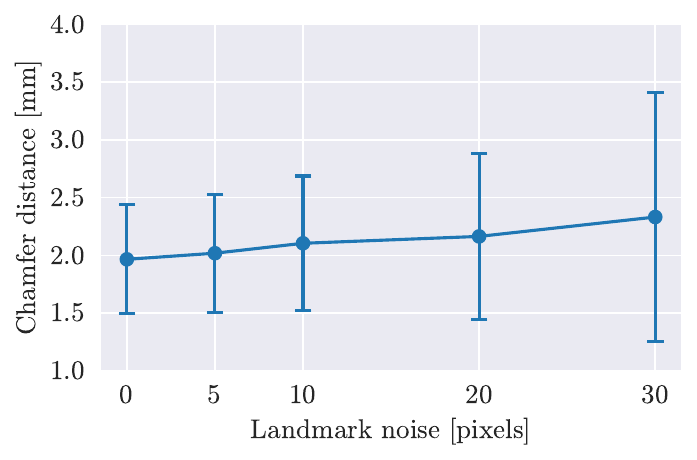}
    \caption{\textbf{Sensitivity to landmark selection.} We test how sensitive our reconstruction pipeline is against noise in the 2D landmark selection process. Our method is quite robust, leaving reconstruction quality nearly unchanged up to an uncertainty of 10 pixels.}
    \label{fig:ablation_landmark_noise}
\end{figure}

\paragraph{Sensitivity to Landmark Selection.}
We test how sensitive our reconstruction pipeline is regarding the selection of the six 2D landmarks required for aligning the SfM-generated point cloud to our model's mean shape.
To this end, we added varying amounts of noise to the ground truth landmark positions (simulated as random perturbations drawn from a discrete uniform distribution in range $\{-k,\dots,-1,0,1,\dots,k\}$ pixels, where $k\in\{5, 10, 20, 30\}$) and run our pipeline.
We repeated this procedure ten times for each noise level and all subjects in our test dataset.
The results in Figure \ref{fig:ablation_landmark_noise} demonstrate that uncertainty in the landmark selection process of up to 10 pixels---corresponding to a spherical uncertainty region with $\sim$5 mm radius in real-world around each landmark---has a negligible impact on reconstruction quality, with CD rising only slightly from 1.97 mm (no noise) to 2.1 mm.
Beyond this point, CD increases steadily, reaching about 2.3 mm at 30 pixels of noise, which corresponds to an uncertainty region with a radius of about 14 mm.
The increasing standard deviation further suggests that individual reconstructions become increasingly susceptible to noise.

\begin{figure}
    \centering
    \includegraphics[width=\linewidth]{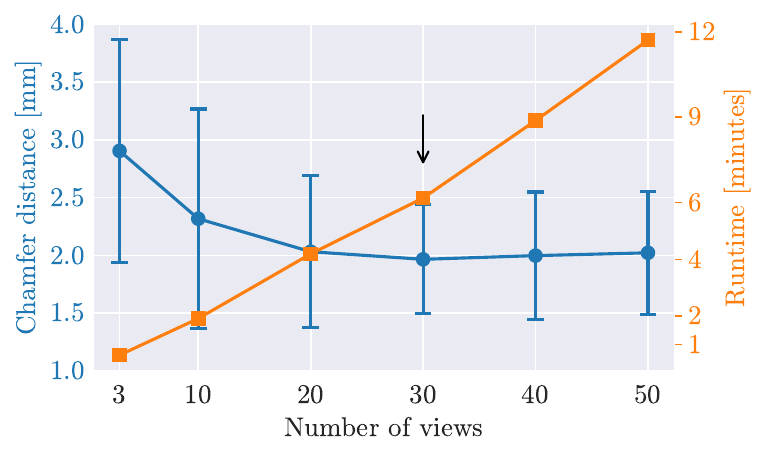}
    \caption{\textbf{Number of views vs. reconstruction quality and runtime.} We ablate the number of frames extracted from the input video and fed into our 3D reconstruction pipeline. We found that, in our setting, 30 images work best, being the ideal trade-off between speed and accuracy.}
    \label{fig:ablation_num_views}
\end{figure}

\paragraph{Number of Views vs. Reconstruction Quality and Runtime.}
We further evaluate how reconstruction accuracy and runtime of our pipeline is affected by the number of views extracted from the input video.
Since our method recovers 3D surfaces by fitting onto point clouds generated by SfM, the number \textit{and} quality of input views directly impact the final 3D reconstruction results.
As shown in Figure \ref{fig:ablation_num_views}, VGGSfM reliably estimates camera poses and produces sufficiently accurate point clouds even with as few as ten input images, achieving a CD of already 2.3 mm when evaluating the reconstructed surface.
Adding more views consistently improves reconstruction quality, peaking at 30 images.
Beyond this point, performance slightly declines, as the inclusion of additional (low-quality) frames into SfM \textit{inevitably} introduces more noise, leading to stagnation or a mild decrease in reconstruction quality.
We also tested our method in a few-shot setting, using only three input frames, matching the number of images required by the commercial Vectra H2 system.
While the reconstruction quality is quantitatively reduced, the error remains below 3 mm, and the visual results are still compelling, see Appendix \ref{app:qual_varying_views}.
Finally, we observe that the runtime---measured only for the SfM step, as this is the primary bottleneck (the rest of the pipeline completes in under a minute)---increases linearly with the number of input images, remaining by about six minutes for 30 images.
Please note that, while VGGSfM is generally reported to be faster \citep{wang2024vggsfm}, our adapted parameter settings lead to slightly longer runtimes.
We refer to Appendix \ref{app:vggsfm_parameter} for a detailed discussion on how these parameters influence reconstruction quality and runtime.

\section{Limitations and Future Work}
While the introduced model and 3D reconstruction pipeline represent a significant step toward making 3D parametric breast models and reconstruction methods more expressive, accurate, and accessible, they still come with certain limitations and opportunities for further improvement.

First, due to the lack of automatic landmark detection algorithms specifically designed for the female breast, our surface reconstruction pipeline currently requires manual annotation of six anatomical landmarks in a single image.
Although this process usually takes no more than five seconds, once such automatic landmarking methods become available, they can be seamlessly integrated into our pipeline, enabling fully automated surface reconstruction.
Furthermore, such landmarking algorithms would not only make our surface reconstruction pipeline easier to use, but also more accurate, as back-projecting 2D landmarks from just one image is less precise than triangulating landmarks selected in \textit{every} image.
However, since no suitable automatic landmarking methods are currently available, manually annotating landmarks in every image would be very time-consuming and infeasible during daily clinical routine.
Therefore, we opted for a practical trade-off between speed, ease of use, and accuracy by requiring manual landmark annotation only once in a single image.
Moreover, it is important to note that our current approach of back-projecting landmarks via closest point searches yields accurate results only if the SfM-generated point cloud is sufficiently dense.
We did not encounter issues in practice, as VGGSfM reliably reconstructs dense point clouds even from low-texture images, typically containing well over 5,000 points.

Second, our model \textit{occasionally} struggles to precisely reconstruct the arms.
This limitation arises from the fact that we did not model arms as articulated body parts, and the presented pose variability in the training data is simply not enough for the model to generalize across a wide range of arm positions.
A potential solution would be to model the thorax as an articulated body as in \citep{palafox2021npm}, for instance.
However, this would require a dataset containing 3D breast scans in a canonical pose, along with additional scans of the same subjects captured in diverse arm positions.
Such data that is currently unavailable and also not easy to acquire in routine clinical practice. 

Finally, in future work, we plan to further evaluate the proposed surface reconstruction pipeline in a medical setting and a larger cohort, following common practice and assessing reconstruction quality in terms of breast volume and anthropometric measurements.

\section{Conclusion}
In this work, we have presented a state-of-the-art neural parametric 3D breast shape model, together with a low-cost and accessible surface reconstruction pipeline that accurately recovers metrically correct 3D breast geometry in minutes from a single monocular RGB video.
Our approach requires neither specialized hardware nor proprietary software and can be used with any device capable of capturing RGB videos.
At the core of our method lies an off-the-shelf differentiable SfM pipeline combined with robust model-based surface reconstruction.
Unlike the existing iRBSM, which models breast geometry using a single \textit{global} neural SDF, our model adopts a space partitioning strategy from state-of-the-art face models: the implicit breast domain is decomposed into multiple smaller regions, each represented by a \textit{local} SDF anchored at anatomical landmark positions.
When integrated into our reconstruction pipeline, the proposed model, dubbed \textit{localized} iRBSM or liRBSM for short, achieves a significantly higher level of detail compared to the iRBSM, precisely recovering breast surfaces within a less than 2 mm error margin, and even capturing fine details such as skin folds and nipples.
Both our model and the surface reconstruction pipeline are publicly available, and we hope this encourages clinicians and other researchers to further develop affordable, accessible, and open-source breast reconstruction methods.
Openly sharing tools and data remains rare in this field, and we aim to help foster greater transparency and collaboration in breast shape modeling and reconstruction research.


\acks{We thank Marc Stamminger for providing valuable ideas and fruitful discussions.
This work was partially funded by the German Federal Ministry of Education and Research (BMBF), FKZ: 01IS22082 (IRRW). 
The authors are responsible for the content of this publication.
The authors gratefully acknowledge the scientific support and HPC resources provided by the Erlangen National High Performance Computing Center (NHR@FAU) of the Friedrich-Alexander-Universität Erlangen-Nürnberg (FAU) under the NHR project b112dc IRRW. NHR funding is provided by federal and Bavarian state authorities. NHR@FAU hardware is partially funded by the German Research Foundation (DFG) – 440719683.}

%
\ethics{The work follows appropriate ethical standards in conducting research and writing the manuscript, following all applicable laws and regulations regarding treatment of animals or human subjects.}

\coi{We declare we don't have conflicts of interest.}

\data{A core contribution of this publication is to share the proposed model, liRBSM, and surface reconstruction pipeline. As such, our trained model (in form of checkpoints containing the optimized weights), along with source code for training and inference as well as code for our 3D surface reconstruction method, is publicly available at \url{https://rbsm.re-mic.de/local-implicit}. Due to the sensitivity of the data and privacy constraints, we can not share the dataset used to train the model nor our test dataset.}

\bibliography{camera-ready}


\clearpage
\appendix

\section{Definition of Metrics}
\label{app:metrics}
We compute metrics between the ground truth 3D breast scan and reconstruction by sampling 100k points and corresponding surface normals from the respective meshes, denoted as $X_\text{gt},N_\text{gt}\subset\mathbb{R}^3$ and $X_\text{rec},N_\text{rec}\subset\mathbb{R}^3$.

\paragraph{Chamfer Distance.}
The Chamfer distance (CD) is defined as 
\begin{equation}
    \frac{1}{2}(\text{Comp.}+\text{Acc.}),
\end{equation}
where
\begin{align}
    \text{Comp.}&=\frac{1}{|X_\text{gt}|}\sum_{x_\text{gt}\in X_\text{gt}}\min_{x_\text{rec}\in X_\text{rec}}\Vert x_\text{gt}-x_\text{rec}\Vert, \\
    \text{Acc.}&=\frac{1}{|X_\text{rec}|}\sum_{x_\text{rec}\in X_\text{rec}}\min_{x_\text{gt}\in X_\text{gt}}\Vert x_\text{rec}-x_\text{gt}\Vert.
\end{align}

\paragraph{F-Score.}
The F-Score is given by
\begin{equation}
    \frac{2\cdot\text{Precision}\cdot\text{Recall}}{\text{Precision}+\text{Recall}},
\end{equation}
where
\begin{align}
    \text{Precision}&=\frac{|\{x_\text{rec}\in X_\text{rec}:\min_{x_\text{gt}\in X_\text{gt}}\Vert x_\text{gt}-x_\text{rec}\Vert<\xi\}|}{|X_\text{rec}|}, \\
    \text{Recall}&=\frac{|\{x_\text{gt}\in X_\text{gt}:\min_{x_\text{rec}\in X_\text{rec}}\Vert x_\text{rec}-x_\text{gt}\Vert<\xi\}|}{|X_\text{gt}|}.
\end{align}
We use $\xi=2.5$ mm for all of our experiments.

\paragraph{Normal Consistency.}
The normal consistency (NC) is computed as
\begin{equation}
    \frac{1}{2}\left(\sum_{x_\text{gt}\in X_\text{gt}}|\langle n_\text{gt},n_\text{rec}(x_\text{gt})\rangle|+\sum_{x_\text{rec}\in X_\text{rec}}|\langle n_\text{rec},n_\text{gt}(x_\text{rec})\rangle|\right),
\end{equation}
where $n_\text{rec}(x_\text{gt})$ denotes the surface normal at the closest point of $x_\text{gt}$ in $X_\text{rec}$.
Similarly, $n_\text{gt}(x_\text{rec})$ denotes the surface normal at the closest point of $x_\text{rec}$ in $X_\text{gt}$.

\section{Additional Results for Intrinsic Model Evaluation}
\label{app:intrinsic_model_additional}
Here, we provide results for the experiment described in Section \ref{subsec:intrisic_evaluation} of the main paper, which investigates the influence of the number of input points during surface fitting on clean point clouds.

\begin{figure}[t]
    \centering
    \includegraphics[width=\linewidth]{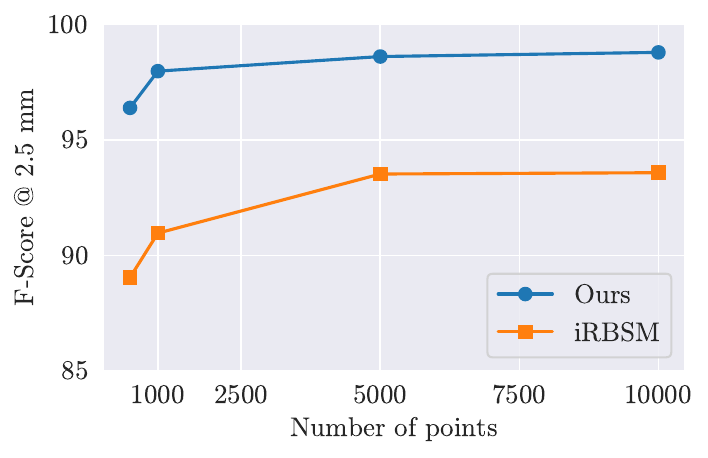}
    \caption{\textbf{Surface fitting under varying point densities.} We investigate the influence of the number of input points used during surface fitting. Our model outperforms the iRBSM across all input densities, with reconstruction quality improving as more points are provided. Higher is better.}
    \label{fig:ablation_num_points}
\end{figure}

\begin{figure}
    \centering
    \includegraphics[width=\linewidth]{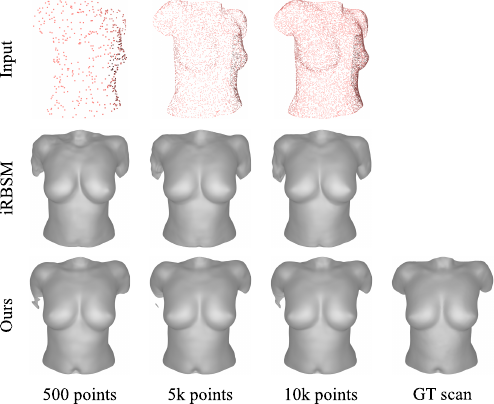}
    \caption{\textbf{Qualitative results for surface fitting under varying point densities.} Comparison of surface reconstructions obtained with the iRBSM and our model when fitted to point clouds of varying densities.}
    \label{fig:ablation_num_points_qualitative}
\end{figure}

Figures \ref{fig:ablation_num_points} and \ref{fig:ablation_num_points_qualitative} show quantitative and qualitative results, respectively. 

\section{Qualitative Results for Ablations}
\label{app:ablation_additional}
\begin{figure*}[t]
    \centering
    \includegraphics[width=\linewidth]{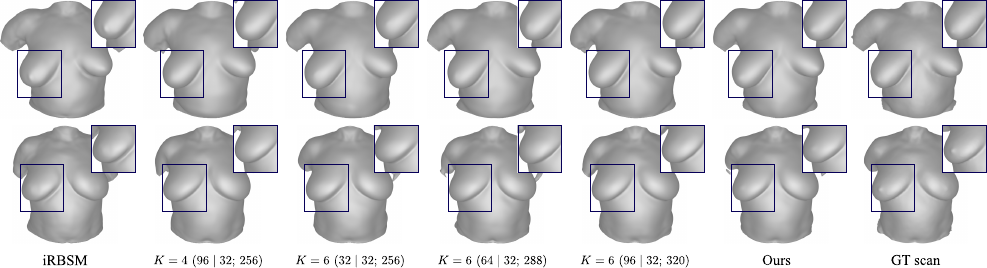}
    \caption{\textbf{Qualitative results for the ablation on anchor layout and latent dimensions.} Again, numbers in parentheses represent global, local, and total latent dimension, \textit{i.e.}, $(L_\text{glob}\mid L_\text{loc};L)$.}
    \label{fig:qual_results_ablation}
\end{figure*}

We provide accompanying qualitative results in Figure \ref{fig:qual_results_ablation} corresponding to the ablation study presented in Section \ref{subsec:ablations} of the main paper.

\section{Qualitative Results for Varying Number of Views}
\label{app:qual_varying_views}
\begin{figure}[h]
    \centering
    \includegraphics[width=\linewidth]{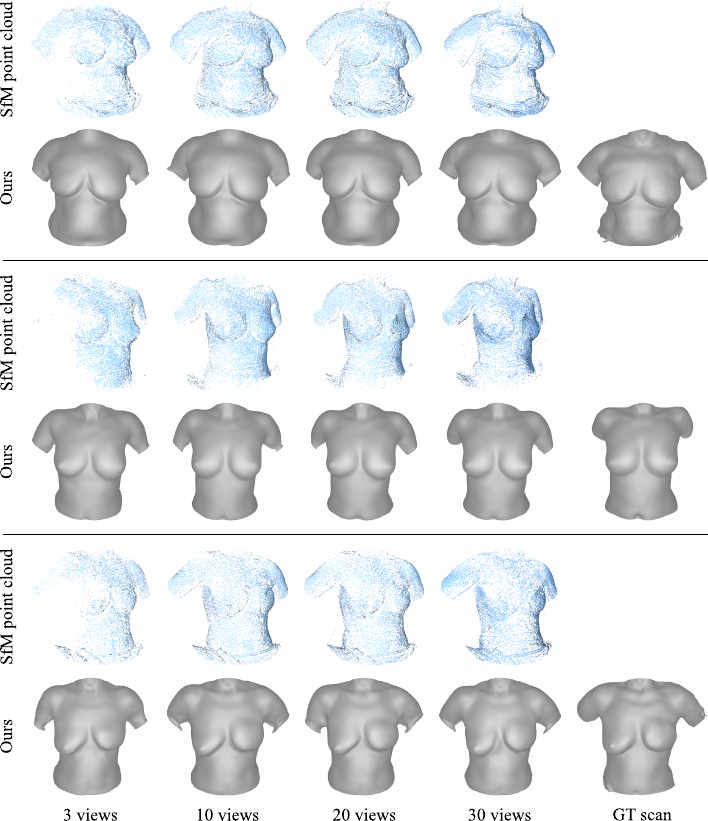}
    \caption{\textbf{Surface reconstruction from monocular RGB videos under varying number of views.} We show surface reconstructions obtained by varying the number of extracted input views to our pipeline.}
    \label{fig:qual_results_num_views}
\end{figure}

We report qualitative results for the ablation study on the number of input views to our 3D surface reconstruction pipeline (see Section \ref{subsubsec:pipeline_ablations} of the main paper) in Figure \ref{fig:qual_results_num_views}.

\section{Discussion On VGGSfM Parameters}
\label{app:vggsfm_parameter}
As briefly mentioned in Section \ref{subsubsec:pipeline_ablations} of the main paper, VGGSfM's runtime is actually reported to be faster \citep{wang2024vggsfm} compared to what we state in our paper, which is due to an adapted parameter.
VGGSfM essentially has two parameters that mainly influence the reconstruction result: \texttt{query\_frame\_num} and \texttt{max\_query\_pts} (per default set to 3 and 2,048).
We found that increasing \texttt{max\_query\_pts} to 8,196 significantly improves surface reconstruction quality.
In Table \ref{tab:ablation_vggsfm}, we provide a small ablation on the two parameters to justify our adaptation.

Generally, we observe that both parameters lead to improved surface reconstructions but longer runtimes when increased up to a certain point. 
Specifically, although a \texttt{query\_frame\_num} of 7 and setting \texttt{max\_query\_pts} to 8,192 leads to the overall best result, due to the long runtime (almost 15 minutes), we decided against using this configuration and chose instead the best trade-off between speed and accuracy, which is the default \texttt{query\_frame\_num} of 3 and \texttt{max\_query\_pts} of 8,192.

\begin{table}[h]
    \centering
    \begin{tabular*}{0.7\linewidth}{l@{\extracolsep\fill}cc}
    \toprule
    & CD $\downarrow$ & Runtime $\downarrow$ \\
    \midrule
    3 $\mid$ 2,048 & 3.06 & 1.56 \\
    3 $\mid$ 4,096 & 2.08 & 3.05 \\
    \underline{3} $\mid$ \underline{8,192} & \underline{1.97} & \underline{6.11} \\
    3 $\mid$ 16,384 & 2.39 & 11.70 \\
    \midrule
    5 $\mid$ 4,096 & 2.07 & 4.96 \\
    5 $\mid$ 8,192 & 1.97 & 10.24 \\
    \midrule
    7 $\mid$ 4,096 & 2.04 & 6.95 \\
    7 $\mid$ 8,192 & 1.93 & 14.47 \\
    \bottomrule
    \end{tabular*}
    \caption{\textbf{Ablation on VGGSfM parameters.} Numbers represent \texttt{query\_frame\_num} and \texttt{max\_query\_pts}. Runtime given in minutes. We \underline{underline} the configuration we have used for our experiments.}
    \label{tab:ablation_vggsfm}
\end{table}

\end{document}